\documentclass[conference]{IEEEtran}

\usepackage{times}
\usepackage{epsfig}
\usepackage{graphicx}
\usepackage{amsmath}
\usepackage{amssymb}
\usepackage{amsfonts}
\usepackage{multirow}
\usepackage{colortbl}
\usepackage{subfigure}
\usepackage{array}
\usepackage{algorithm}
\usepackage{hhline}
\usepackage{color}
\usepackage{xspace}
\usepackage{ifpdf}
\usepackage{booktabs}
\usepackage{url}

\usepackage[numbers]{natbib}
\usepackage{multicol}
\usepackage[bookmarks=true]{hyperref}

\def\ie{\emph{i.e.~}}

\def\footnoterule{\relax%
	\kern-5pt
	\hbox to \columnwidth{\vrule width 0.5\columnwidth height 0.4pt\hfill}
	\kern4.6pt}
\makeatother

\IEEEoverridecommandlockouts

\graphicspath{{./figs/}}

\begin{document}
	
	\title{\LARGE \bf SE-SLAM: Semi-Dense Structured Edge-Based Monocular SLAM}
	
	\author{Juan Jos{\'e} Tarrio$^{1,2,3}$, Claus Smitt$^{1,2}$, Sol Pedre$^{1,2}$\\
		\thanks{All work made exclusively while the authors were working at Instituto Balseiro, Comision Nacional de Energia Atomica (CNEA)}
		
		\small
		$^{1}$ Instituto Balseiro
		$^{2}$ CNEA 
		$^{3}$ SLAMCore ltd.		
	}
	
	\maketitle

\begin{abstract}

Vision-based Simultaneous Localization And Mapping (VSLAM) is a mature problem in Robotics. Most VSLAM systems are feature based methods, which are robust and present high accuracy, but yield sparse maps with limited application for further navigation tasks. Most recently, direct methods which operate directly on image intensity have been introduced, capable of reconstructing richer maps at the cost of higher processing power.  
In this work, an edge-based monocular SLAM system (SE-SLAM) is proposed as a middle point: edges present good localization as point features, while enabling a structural semi-dense map reconstruction. However, edges are not easy to associate, track and optimize over time, as they lack descriptors and biunivocal correspondence, unlike point features. To tackle these issues, this paper presents a method to match edges between frames in a consistent manner; 
a feasible strategy to solve the optimization problem, since its size rapidly increases when working with edges; and the use of non-linear optimization techniques. 
The resulting system achieves comparable precision to state of the art feature-based and dense/semi-dense systems, while inherently building a structural semi-dense reconstruction of the environment, providing relevant structure data for further navigation algorithms. 
To achieve such accuracy, state of the art non-linear optimization is needed, over a continuous feed of $~10000$ edgepoints per frame, to optimize the full semi-dense output. Despite its heavy processing requirements, the system achieves near to real-time operation, thanks to a custom built solver and parallelization of its key stages. In order to encourage further development of edge-based SLAM systems, SE-SLAM source code will be released as open source.


%
%

\end{abstract}

\section{Introduction}

Vision-based Simultaneous Localization And Mapping (VSLAM) is a well studied problem, and key for Computer Vision and Robotics. SLAM lies at the core of most systems that require awareness of self motion and scene geometry.

Many solutions have been proposed for the VSLAM problem. Feature based algorithms \cite{Mur2015tor,Mur2017tor,Leutenegger2015ijjr, Bloesch2017ijrr, Forster2017tor, Qin2018TOR} rely on recognition and tracking of point landmarks over time. These methods are robust and present high accuracy, however the reconstructed map is sparse, thus presents little practical use by itself. They also present problems in textureless environments, where point features are difficult to come by \cite{Ojeda2017pl}, \cite{Pumarola2017pl}.   

Direct methods operate directly on image intensity, reconstructing depth at every image pixel \cite{Newcombe2011iccv,Forster2014icra} or in a selected subset of pixels \cite{Engel2018pami}. These methods avoid the cost of feature extraction, but state of the art methods need to estimate and correct a photometric camera model to deal with illumination changes and occlusions, 
yielding a more complex optimization problem \cite{Engel2018pami}. Nevertheless, the reconstruction output of these systems usually yields richer maps, more suitable for navigation than feature-based methods, at the cost of higher processing power and GPU processing in full dense methods\cite{Newcombe2011iccv}. 

In this context, an edge-based system appears as an interesting middle point between the former. On one hand edge extraction is robust, highly parallelizable and widely studied in computer vision, and it is also known for taking part in biological vision systems. Moreover, edges are trackable features
, meaning that the geometric re-projection error can be employed as done in classical feature-based systems. Furthermore, edges can be extracted in textureless scenes where classical feature extraction is difficult. Finally, an edgemap usually contains objects boundaries, hence its reconstruction provides rich information about structure of the observed scene, and may play a part in full dense reconstruction and semantic interpretation.
\begin{figure}[t!]
	\centering
	\resizebox{0.9\columnwidth}{!}{\includegraphics{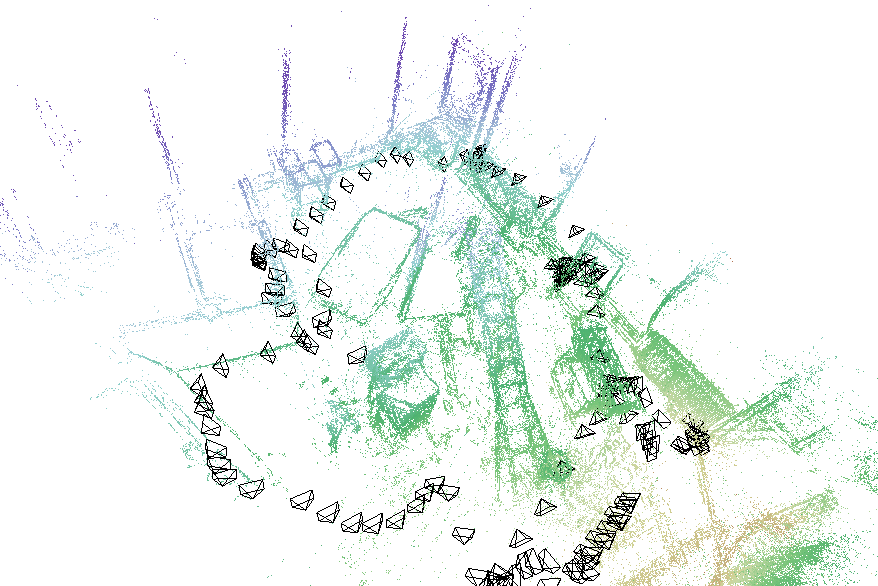}}
	\caption{Final edgemap and trajectory obtained in the EuRoC Vicon Room 2 dataset}
	\label{fig:reconstruction}
\end{figure}

Despite its potential benefits, SLAM systems using edges are not common and most of them rely on straight lines as a complement to classical features, to increase robustness when base methods fail \cite{Smith2006bmva, Pumarola2017pl,Ojeda2017pl}. Only recently an edge-based system has been proposed in the literature \cite{Maity2017iccv}. This is probably due to the fact that edges are not easy to associate, track and optimize over time, as they lack descriptors and do not present biunivocal correspondence, unlike point features.

In this work, a novel edge-based SLAM system is introduced (SE-SLAM), which relies on edge detection as sole input, except for the loop closure step. 
Tracking and data association is tackled by following up on previous developments in edge-based visual odometry for MAVs control \cite{Jose2015iccv,Tarrio2018jint}. Full non-linear keyframe based optimization is performed on edges, adapting recent methods \cite{Leutenegger2015ijjr,Engel2018pami} to the limitations and advantages presented by this type of feature. The resulting system achieves comparable accuracy to state of the art methods and provides a full edge reconstruction output, suitable for further navigation steps (see Figure \ref{fig:reconstruction}) \cite{Von2017ecmr}. Using this type of back-end to optimize a pure edge input is, to the authors' knowledge, the first of its kind. 

\section{Related Work}


Feature based algorithms that rely on recognition and tracking of point landmarks over time are the most common solution to the Visual SLAM problem \cite{Mur2015tor,Mur2017tor,Leutenegger2015ijjr}. 
ORB-SLAM2 \cite{Mur2017tor} is probably the state of the art in terms of accuracy. OKVIS \cite{Leutenegger2015ijjr} is another relevant solution, targeted to stereo visual-inertial odometry, whose optimization and marginalization approach presented served as inspiration for this work. These methods present high accuracy, however the reconstructed map is sparse and a separate mapping algorithm is required to get denser ones \cite{Mur2015ss}. 


Addressing this issue, Direct SLAM methods deal with the problem of minimizing an intensity based functional in all image pixels without a feature extraction phase. DTAM \cite{Newcombe2011iccv} was the first to show a dense 3D monocular reconstruction of this kind. It relies on heavy GPU aided processing to incrementally fuse and regularize every image pixel into a dense model and is constrained to small environments. 
As an effort to reduce the computational cost of these algorithms, semi-dense approaches were presented \cite{Engel2014eccv} that operate over image intensities but only on a meaningful set of pixels. These pixels are picked based on their traceability, using the high intensity gradient parts of the image. Hence, its output contains edges but no explicit treatment is made. DSO \cite{Engel2018pami} applies non-linear optimization to an intensity based functional over a sparse selection of pixels. Moreover, they incorporate lighting parameters to the camera model to account for illumination and exposition time changes, making the optimization more complex. In addition, dealing with individual pixels makes association even harder compared to edges. In \cite{Engel2018pami} this is tackled by redefining variables on each keyframe, which leads to variable and measurement repetition, consequently reducing performance. 
In the present work, illumination changes are handled by edge detection, and association is used to reduce the number of repeated variables, enabling whole edgemap reconstruction with non-linear optimization. 

Furthermore, SLAM and visual odometry systems explicitly using edges have also been introduced. The ones using edges as sole input, actually use fit straight lines \cite{Zhang2011icra, Zhou2015tvt, Zhang2015TRO, Tomono2014icra}, hence their application is restricted to structured scenarios where these features are present. 
On the other hand, line features have also been used to enhance existing systems, with the aim of boosting their performance in textureless environments, while yielding a structured line map as a secondary goal. A pioneer in this field is \cite{Smith2006bmva}, where authors incorporate lines into a monocular Extended Kalman Filter system. More recently,  \cite{Yang2017} and \cite{PLSVO2016} extended the semi-direct monocular visual odometry system SVO \cite{Forster2014icra} using lines to improve its performance in textureless enviroments. In PL-SLAM \cite{Pumarola2017pl} authors extend ORB-SLAM with lines, presenting a monocular SLAM system that shows improvement in less textured sequences at the cost of nearly duplicating the computational cost of plain ORB-SLAM. Almost at the same time, another PL-SLAM \cite{Ojeda2017pl} that builds upon the ORB-SLAM pipeline was presented, although using a stereo approach. This work presents results showing better performance than ORB-SLAM on specially recorded real-world textureless scenes, without such a high computational burden. They also develop a method for enhanced loop closure using line descriptors, which could be an interesting add-on to the system in this paper. All of these approaches use the LSD line detector \cite{Gioi2010LSDAF} and LBD line descriptor for matching purposes \cite{Zhang2013AnEA}. 

There is not much work in explicitely using general edges in SLAM systems, probably since edges are not easy to associate, track and optimize over time, as they lack descriptors and do not present biunivocal correspondence. Eade and Drummond \cite{Eade2006bmvc} present what may be the first work in this line. They define an edge features called \textit{edgelet} - a very short, locally straight segment of what may be a longer, possibly curved, line - and use it to enhance a particle filter SLAM system, to take advantage of the structural information provided by edges. Klein and Murray \cite{klein2008eccv} also use this \textit{edgelet} definition, but in this case to enhance robustness against rapid camera motions of their PTAM system \cite{Klein2007ismar}. Closest to the present work is the Edge SLAM System presented recently in \cite{Maity2017iccv}, although their optical flow for tracking approach to track edges is significantly different from SE-SLAM. To compare against this last system, section includes results for ICL-NUIM sequences reported by them, showing that SE-SLAM outperforms theirs in all but one sequence. Regrettably, there is no public available code, neither parametrization of their work to compare in more challenging datasets like EuRoC or TUM-VI. 

\section{Contribution}

This paper introduces SE-SLAM, a novel semi-dense structured edge-based monocular SLAM system. 
SE-SLAM shows comparable precision to state of the art feature-based and dense/semi-dense systems, but using edges as sole input in all the stages but loop closure, achieving real-time operation in several cases. The edge-based nature of the system not only helps in textureless environments, but also inherently builds a structural semi-dense reconstruction of the environment suitable for further navigation algorithms. This is accomplished thanks to the following contributions:

\begin{itemize}
\item A method to associate edges between frames and relatively far keyframes, in a consistent manner, that enables the application of techniques traditionally reserved to point features. 
\item A parametrization, residual and variable selection and marginalization strategy that exploits edges' nature, to make the optimization problem treatable, since its size increases rapidly if applying existing methods to edges. 
\item A non-linear optimization approach that better adapts to an edge based input. 
\end{itemize}

The rest of this paper is organized as follows: section~\ref{sec:formulation} presents the overall problem, sections~\ref{sec:tracking}, ~\ref{sec:data_asoc}~and~\ref{sec:kf_sel}  describe the front-end of the system, namely how the initial conditions and data association are generated to build the optimization problem. Section~\ref{sec:backend} presents the back-end and which assumptions and approximations are made to solve the optimization efficiently. Section~\ref{sec:loop_closure} presents a loop closure detection and integration strategy. Finally, some relevant implementation details are explained in section~\ref{sec:imp_details} and results obtained by running the algorithm on public datasets are presented in section~\ref{sec:results}. 



\section{Problem Formulation}
\label{sec:formulation}
The input of the algorithm is composed by edgemaps \cite{Jose2015iccv} defined as sets of edge belonging pixels (edgepoints), including its sub-pixel position $\mathbf{q}$, a measure of local edge gradient or normal direction $\mathbf{m}$ and connections to its adjacent edgepoints. The later play a key role in the data association phase.

The purpose of the system is to recover 3D position of each edgepoint, while at the same time infer camera poses. What makes this problem challenging is that there is no obvious way of differentiating neighbour pixels. Therefore, edgepoints cannot be univocally matched over time, as a matter of fact, the same edge observed from different scales may have different amounts of edgepoints. For this reason, a keyframe-based approach similar to previous semi-dense methods is followed \cite{Engel2014eccv}, edgemaps are defined for each keyframe and anchored to them, and the global map results from the union of all edgemaps. 

However, redefining and optimizing an edgemap per keyframe inevitability leads to variable repetition and a waste of computational resources. To solve this, two types of keyframes will be differentiated thought all this work. Keyframes in which an edgemap is defined are called ``Prior KeyFrames" (PKFs). This corresponds with the fact that for each one of them there will be a measurement prior, or factor added to the pose graph optimization step. On the other hand, and for the sake of clarity, keyframes that only provide edge position measurements will be called ``Data KeyFrames" (DKFs).

Another advantage of this separation is that it allows to use different criteria for keyframe selection, while PKFs are selected based on covisibility (number of novel edgepoints), DKFs are selected using simple heuristics that maximize inter keyframe translation. As done in recent SLAM methods \cite{Mur2015tor}, every new frame is added as a keyframe and later removed from an optimization window according to selection criteria.

\textbf{Notation convention:} PKFs will be indexed with the letter $k$, DKFs with $j$ and edgepoints with $i$.

\subsection{Edge Parametrization}

\begin{figure}[ht!]
	\centering
	\resizebox{0.6\columnwidth}{!}{\includegraphics{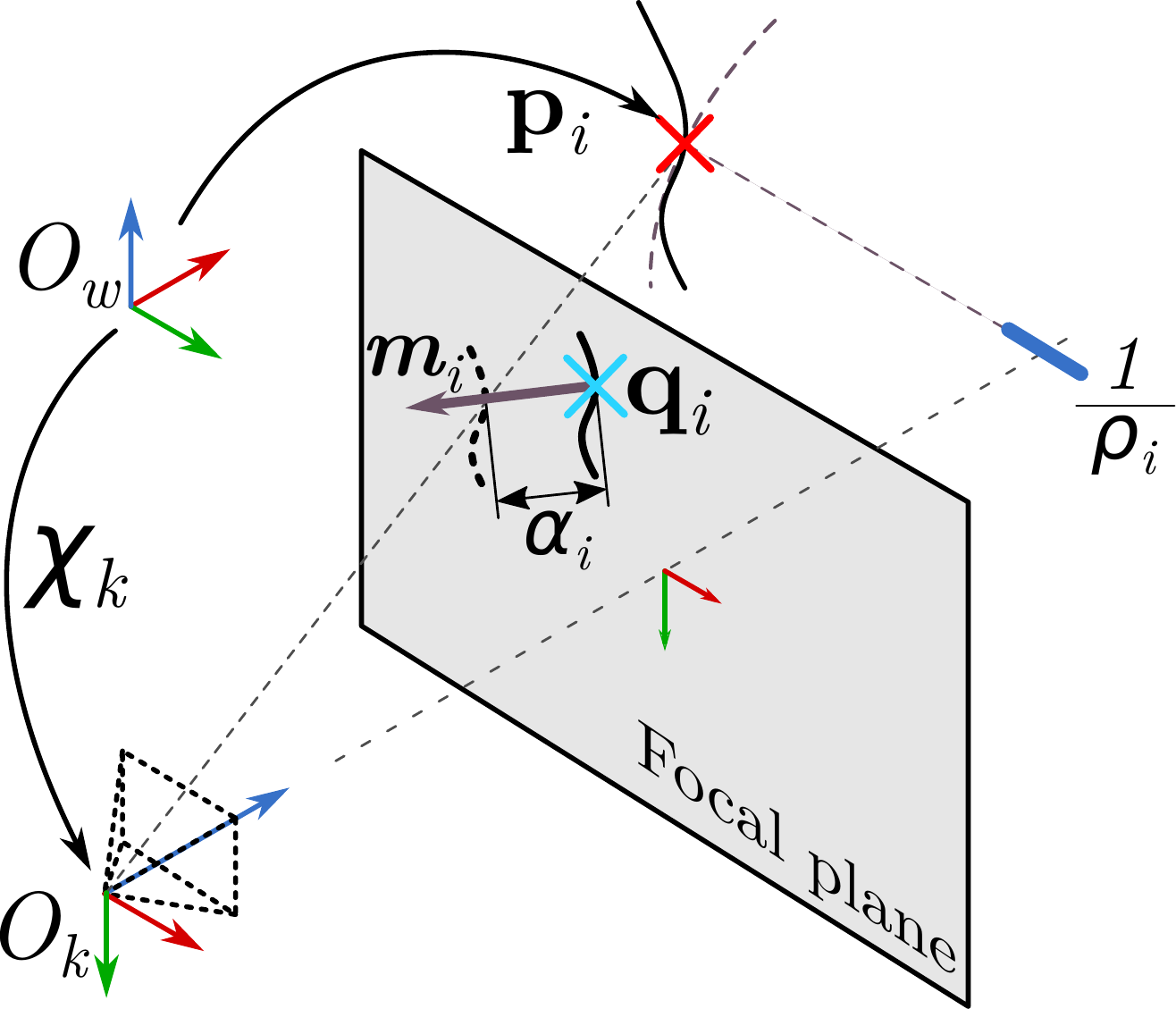}}
	\caption{Edgepoint's parametrization used for tracking and optimization. An image edgepoint measurement $\mathbf{q_i}$ is parametrized with inverse depth $\rho$ and a displacement $\alpha$ along the edge perpendicular direction $\mathbf{m_i}$. Optimizing $\alpha$ accounts for the measurement model of $\mathbf{q_i}$.}
	\label{fig:params}
\end{figure}

In order to perform bundle adjustment, a parametrization must be chosen for spatial landmarks, edgepoints in this case. To account for the fact that edgepoints are defined relative to PKFs, a relative inverse depth parametrization is adopted \cite{Civera2008tog}. However, given that edgepoints present no localization information along the edge direction\cite{Jose2015iccv}, only 2 parameters are selected: inverse depth $\rho$, and distance along the edge's normal direction $\alpha$. This parametrization is sketched in Figure~\ref{fig:params}.

Defining $\mathbf{q}^k_i$ as the image position of edgepoint $i$ in PKF $k$, $\mathbf{m}^k_i$ its normal direction vector and $\pi^{-1}(\mathbf{q}, \rho)$ the back projection from image to euclidean space, the 3D position of the edgepoint in camera coordinates is defined as:
\begin{align}
\mathbf{p_i}= \pi^{-1} (\mathbf{q}^k_i + \mathbf{m}^k_i \alpha_i, \rho_i)
\end{align}

This parametrization constrains the edgepoint to only move along edge's perpendicular direction  during optimization. Hence dealing with uncertainty in localization of edgepoints along edge's tangential direction.
 

\subsection{Error Model}

The error model used by \cite{Jose2015iccv} is kept, which is a reprojection error function that takes into account edge's matching uncertainty along its tangential direction. Hence, only the error along its normal direction is taken into account. For a generic edgepoint $i$ defined in PKF $k$ and associated in DKF $j$, the measurement takes the form:

\begin{align}
\mathbf{e^{k,j}_i}&=\nonumber\\
&\left[\mathbf{q}_i^j - \pi\left(T^{-1}\left(\mathbf{\chi}_j ,T\left(\mathbf{\chi}_k, \pi^{-1} \left(\mathbf{q}^k_i + \mathbf{m}^k_i \alpha_i, \rho_i\right)\right)\right)\right)\right] \bullet \mathbf{m}_i^j \label{eq:rep_error}
\end{align}

Where $\mathbf{\chi}_k, \mathbf{\chi}_j \in SE3$ are the respective keyframe's poses, $\mathbf{q}_i^j$ and $\mathbf{m}_i^j$ are the associated edgepoint's position and normal direction in DKF $j$, and the scalar product yields the projected error along this direction. 

\subsection{Uncertainty and Robust Estimation}

Given a model for edge position uncertainty $\sigma_i^{j}$ provided by the detection method, equation \ref{eq:rep_error} can be modified to account for it. Moreover, IRLS is used in every optimization step weighting the reprojection error to account for outliers. Overall, the residual function takes the form:  

\begin{align}
\mathbf{f^{k,j}_i}&= \frac{w}{\sigma_i^{j}}\mathbf{e^{k,j}_i}
& w=\begin{cases}
1 & \left|e^{k,j}_i\right| < \epsilon\\\frac{1}{|e^{k,j}|} & aoc
\end{cases}\label{eq:residual}
\end{align}

Employing a slightly modified version of Huber weighting. 


To take into account edgepoint position in its defining PKF, an extra measurement must be added as prior so as to constrain $\alpha$, which is not robustly weighted:

\begin{align}
\mathbf{f^{k,k}_i}&= \frac{\alpha_i}{\sigma_i^{k}} \label{eq:residual_prior}
\end{align}

\subsection{Total Energy Functional}

Having defined the error model, the optimization problem can be posed as the minimization of the reprojection errors of edgepoints defined in each PKF with respect to matches in DKFs: 

\begin{align}
E_{total}  = \sum_k E_k = \sum_k \sum_j \sum_i (\mathbf{f^{k,j}_i})^2 \label{eq:energy_total}
\end{align}

Where $E_k$ groups all measurements related to the edgemap defined in PKF $k$ and  $\mathbf{f^{k,j}_i}$ takes the form of equation~\ref{eq:residual} if $k \ne j$ and equation~\ref{eq:residual_prior} otherwise.

\section{Edge Tracking}
\label{sec:tracking}

In the context of this work, tracking is the problem of finding the best transformation between the current and next frame, without prior data association or initial condition. For this purpose the method presented in \cite{Jose2015iccv} is used, with some enhancements to make it more robust against fast motions. 

In \cite{Jose2015iccv} an auxiliary image is created from the new frame which contains in each pixel the index of the closest edgepoint. This image is used to re-project the current frame into the new one and find an initial match for each edgepoint. Later, the re-projection error functional in equation \ref{eq:rep_error} is minimized, only with respect to the relative translation between frames. This is done iteratively, rebuilding the matches in each iteration, and using the current frame's estimated depth as an input datum.

In favor of rejecting some wrong association, a threshold is applied to the difference between gradient vectors of each edgepoint. Optionally, a threshold can also be applied to the intensity difference at both sides of the edge (one comparison is made on each side).

\subsection{Multi Scale Tracking}

The input image is decimated at different scales and an edgemap is extracted for each of them. The tracking algorithm is applied in sequence from coarse to fine scales.

No depth estimation is made for the coarsest scale, instead depth from the main (finest) scale is recursively propagated to the coarser ones using a nearest neighbour approach. Edges that fail to find a close neighbour in the finer scale are rejected. Beside saving running time, this approach chooses coarse edges that have a correspondence in finer scales. This edgepoints usually belong to better localized object boundaries.

\subsection{Depth Smoothing for Tracking and Association}

Only for the tracking stage, edgepoint's depth is smoothed, where each edgepoint is averaged using it's closest neighbours' depth. 

By smoothing depth, a uniform warping of close edges is achieved, thus improving association in image parts with close similar edges. After the tracking stage, smoothed depth is discarded.

\section{Data asociation}
\label{sec:data_asoc}
\begin{figure}[t!]
	\centering
	\resizebox{\columnwidth}{!}{\includegraphics{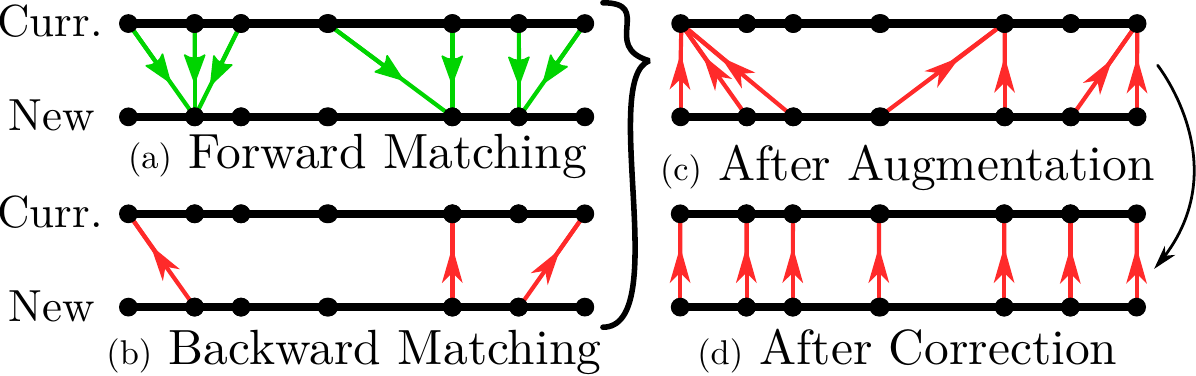}}
	\caption{Edgepoints association-correction process. When backward matching is generated from forward matching, holes are created due to non-biunivocal association. This is corrected first by augmentation and then correction steps.}
	\label{fig:aug_corr}
\end{figure}

The main output of the tracking stage is the optimal transformation between the current and new frame, given the edgemap's structure. However, associations are made from the current edgemap points to the new ones (Fig. \ref{fig:aug_corr}a). Due to the nature of edges, these matches are not biunivocal, as two different edgepoints may be matched with the same edgepoint in the new frame and there is no guarantee that every new edgepoint will be matched (Fig. \ref{fig:aug_corr}b). 

Given how the optimization problem is posed in the back-end, it is important to maximize and refine associations from the new frame to the current one. This is done in two steps: augmentation and correction. Both steps exploit the continuous nature of edges \ie most edgepoint are connected to adjacent ones.

\subsection{Matching augmentation}
In this step each matched edgepoint shares this association with its neighbours. This is done by ``walking'' the edge, which means recursively go over its neighbours and assign the same associated edgepoint until a valid association is found (Fig. \ref{fig:aug_corr}c). 

\subsection{Matching correction} 
For a given edgepoint with an association in a target frame, correction means finding the best matching edgepoint in terms of the epipolar constraint between the two frames obtained by tracking stage estimated pose. For this search the edge is ``walked'' again using the connectivity information. In practice not only the epipolar line is taken into account, also current edgepoint reprojected position help improve cases where the epipolar is closed to parallel with the edge or there is no motion. 

This process is applied to each new frame after augmentation, and is also used in the optimization back-end when building the Jacobians. Both augmentation and correction stages are illustrated in Figure~\ref{fig:aug_corr}.

\section{Variable and Keyframe Selection}
\label{sec:kf_sel}

In the presented system edgemaps are only defined in certain keyframes. Even though it would be possible to propagate associations for every PKF in both directions (past and future frames) and solve a joint problem as done in \cite{Engel2018pami}, this would lead to variables and measurements repetition. 

Ideally, an edgepoint representing an individual edge fragment should be defined as a variable in only one PKF, however this is not easy to achieve due to the non-biunivocal association of edgepoints. Nevertheless an approximation may be obtained by using the associations to disable redefined edgepoints.

An open question is in which particular PKF should an edgepoint be defined. Possible options are to anchor the edgepoint to the first or last PKF where it was observed. Even though the former has its advantages, the later is chosen as it assures that all currently visible edgepoints are anchored and optimized relative to the last frame. This particular choice keeps depth information ready for the tracking step and it's key for defining the marginalization strategy. In fact, having associations only to the past, assures that keyframe marginalization involves only previous point measurements, therefore closer to the optimal.

When a new frame arrives it's added as PKF to the backend. Associations are back-propagated to match every edgepoint against active PKFs and DKFs. When the amount of edgepoints matched in a particular keyframe falls bellow a fraction of all edgepoints ($\approx10\%$), the association process is stopped and a local covisivility window $S_k$ is defined for PKF $k$. At the same time, the new frame it's kept as PKF if the amount of newly observed edgepoints with respect to the last PKF exceeds a certain threshold, otherwise it's added as DKF.

Every association from new PKFs to the previous one disables the matched variables in the later one, thus reducing variable repetition and computational cost. Ideally, current visible edgepoints would be active variables only in the new PKF, while the others contain only edgepoints that are not visible in the newer ones. In practice, variable repetition remains due to the non-biunivocal matching of edges. However this proves to be beneficial to accuracy as it maximizes the number of measurements used.

When the amount of active keyframes inside the newest covisivility window $S_k$ exceeds a limit (10-15), the DKF that posses the less translation with respect to its consecutive keyframes is discarded. PKFs are not discarded, as they define the map.

\section{Hierarchical Optimization}
\label{sec:backend}
The presented system exploits primary and secondary sparseness \cite{Triggs1999iwva} through a custom built gauss newton solver designed for the posed problem structure. In fact, local co-visibility windows defined for each PKF suggests a natural split of the problem. 

Overall the method goes as follows. First each individual sub-functional $E_k$ is linearised around the current optimization point and edgepoint variables are marginalized using the Schur complement. The resulting priors, involving poses only, are jointly optimized using Pose-Graph Optimization (PGO) over absolute poses. Finally, edgepoints are updated by the linear increments obtained from the PGO.
\subsection{Edgepoint Marginalization}
\label{sec:schur}
Instead of posing the problem in absolute coordinates like \cite{Engel2014eccv} and \cite{Leutenegger2015ijjr}, a mixed absolute - relative formulation is chosen, a reasoning for this is given in section \ref{sec:rel_coords}. For a given sub-functional $E_k$, residuals are expressed in terms of the relative transformation that takes edgepoints from the PKF $k$ to the corresponding keyframe $j$:
\begin{align}
r^{k,j}_i&=\frac{w}{\sigma^j_i}
\left[\mathbf{q}_i^j - \pi\left(T\left(\mathbf{\chi}_k^j, \pi^{-1} \left(\mathbf{q}^k_i + \mathbf{m}^k_i \alpha_i, \rho_i\right)\right)\right)\right] \bullet \mathbf{m}_i^j \label{eq:rep_error_rel}
\end{align}
Where $\mathbf{\chi}_k^j$ is a transformation belonging to SE3 Group defined by the composition function:
\begin{align}
\mathbf{\chi}_k^j = \mathcal{C}\left(\mathbf{\chi}_j, \mathbf{\chi}_k\right) = \mathbf{\chi}_j^{-1} \mathbf{\chi}_k = \left[ \begin{array}{cc}
R^j_k&\mathbf{t^j_k} \\0 & 1
\end{array}\right] \label{eq:comp_func}
\end{align}
Considering that projection function $\pi(\mathbf{p})$ is invariant to scale, eq.\ref{eq:rep_error_rel} can be re-written as:
\begin{align}
r^{k,j}_i&=\frac{w}{\sigma^j_i}
\left[\mathbf{q}_i^j - \pi\left( R^j_k (\mathbf{q_h}_i^k + \mathbf{m_h}_i^k \alpha_i) \rho_i^{-1} + \mathbf{t^j_k}\right)\right] \bullet \mathbf{m}_i^j \\
&=\frac{w}{\sigma^j_i}
\left[\mathbf{q}_i^j - \pi\left( R^j_k (\mathbf{q_h}_i^k + \mathbf{m_h}_i^k \alpha_i) + \mathbf{t^j_k} \rho_i\right)\right] \bullet \mathbf{m}_i^j \label{eq:rep_error_rel_2}
\end{align}
Where $\mathbf{q_h}$, $\mathbf{m_h}$ and $\pi(\mathbf{p})$ are defined as:
\begin{align}
\mathbf{q_h}&= \left[ \begin{array}{c}
(q_{x} - c_x) f_x^{-1} \\ (q_y - c_y) f_y^{-1}\\ 1
\end{array}\right] ; \mathbf{m_h}= \left[ \begin{array}{c}
m_{x} f_x^{-1} \\m_y f_y^{-1}\\ 0
\end{array}\right] \\
\pi(\mathbf{p})&= \left[ \begin{array}{c}
p_x p_z^{-1} f_x + c_x \\ p_y p_z^{-1} f_y + c_y
\end{array}\right]
\end{align}
To take into account edgepoints position measurement in the respective PKF, a prior on $\alpha$ is added:
\begin{align}
\mathbf{r^{k,k}_i}&= \frac{\alpha_i}{\sigma_i^{k}} \label{eq:residual_prior_2}
\end{align}

Throught all optimization, analytical expressions for the residual~\ref{eq:rep_error_rel_2} with respect to $\rho$, $\alpha$ and transformations are used:
\begin{align}
J^{k,j}_{i,\rho} &=\frac{\partial r^{k,j}_i}{\partial \rho_i}= 
\frac{w}{\sigma^j_i} J_{\pi \bullet \mathbf{m}_i^j} \mathbf{t^j_k}\\
J^{k,j}_{i,\alpha} &=\frac{\partial r^{k,j}_i}{\partial \rho_i}= 
\frac{w}{\sigma^j_i} J_{\pi \bullet \mathbf{m}_i^j} R^j_k \mathbf{m_h}_i^k\\
J_{\pi \bullet \mathbf{m}_i^j}&=\left. \frac{\partial \pi(\mathbf{p}) \bullet \mathbf{m}_i^j}{\partial \mathbf{p}}\right|_{\mathbf{p}= R^j_k (\mathbf{q_h}_i^k + \mathbf{m_h}_i^k \alpha_i) + \mathbf{t^j_k} \rho_i}
\end{align}
Jacobians over SE3 elements are taken with respect to an infinitesimal increment in SE3 tangent space $\delta \mathbf{\chi}_k^j = [\mathbf{v}_k^j ; \mathbf{\omega}_k^j]$ (refer to \cite{Blanco2010tr}, \cite{Barfoot2017CUP}) taking into account equation~\ref{eq:rep_error_rel_2}, it can be shown that:
\begin{align}
J^{k,j}_{i,\mathbf{v}_k^j} &= 
\frac{w}{\sigma^j_i} J_{\pi \bullet \mathbf{m}_i^j} \rho_i \\
J^{k,j}_{i,\mathbf{\omega}_k^j} &= 
\frac{w}{\sigma^j_i} J_{\pi \bullet \mathbf{m}_i^j} \left[R^j_k (\mathbf{q_h}_i^k + \mathbf{m_h}_i^k \alpha_i) + \mathbf{t^j_k} \rho_i\right]_\times
\end{align}
With these Jacobians and stacking residuals and variables into vectors, it is possible to obtain a second order approximation of the functional $E_k$ that takes the known form:
\begin{align}
E_k &\left(\mathbf{\delta x}_k \boxplus \mathbf{\bar{x}}_k\right) \simeq
E^0_k+  
2\mathbf{g}^{k^T}\mathbf{\delta x}_k 
+\mathbf{\delta x}_k^T H^k \mathbf{\delta x}_k \label{ec:shur0}
\end{align}
Whose vector of variables is defined as:
\begin{align*}
\mathbf{x}_k &= \left[\begin{array}{ccc}
 \mathbf{\chi}^{k-1}_k \dots \mathbf{\chi}^{k-S_k}_k 
&\alpha^k_1 \dots \alpha^k_{N^k} 
&\rho^k_1 \dots \rho^k_{N^k} \end{array}\right]\\
&\in SE3^{S_k} \times R^{N^k} \times R^{N^k} 
\end{align*}
Being $S_k$ the $k^{th}$ frame co-visibility window length, and its increments are:
\begin{align*}
\mathbf{\delta x} &= \left[\begin{array}{ccc}
\mathbf{\delta \chi}^{k-1}_k \dots \mathbf{\delta \chi}^{k-S_k}_k 
&\delta \alpha^k_1 \dots \delta \alpha^k_{N^k} 
&\delta \rho^k_1 \dots \delta \rho^k_{N^k} \end{array}\right]\\
&= \left[\begin{array}{ccc}
\mathbf{\delta \chi}^k
&\mathbf{\delta \alpha}^k
&\mathbf{\delta \rho}^k \end{array}\right]\\
&\in R^{6 ^ {S_k}} \times R^{N^k} \times R^{N^k} 
\end{align*}
Functional \ref{ec:shur0} has a known form in bundle adjustment \cite{Triggs1999iwva} that enables its components to be defined as:
\begin{align}
E_0^k(\mathbf{\bar{x}}) &= \mathbf{r_k}^{T}\mathbf{r}_{k}\\
\mathbf{g}^{k^T}(\mathbf{\bar{x}}) &= \mathbf{r_k}^{T}\left[\begin{array}{ccc} J^{k}_{\mathbf{\chi}}& J^{k}_{\mathbf{\alpha}} &J^{k}_{\mathbf{\rho}} \end{array}\right] 
\\
H^k(\mathbf{\bar{x}}) &=
\left[\begin{array}{ccc} 
H_{\mathbf{\chi},\mathbf{\chi}}^k&H_{\mathbf{\chi},\mathbf{\alpha}}^k &H_{\mathbf{\chi},\mathbf{\rho}}^k\\
H_{\mathbf{\alpha},\mathbf{\chi}}^k&H_{\mathbf{\alpha},\mathbf{\alpha}}^k &H_{\mathbf{\alpha},\mathbf{\rho}}^k\\
H_{\mathbf{\rho},\mathbf{\chi}}^k&H_{\mathbf{\rho},\mathbf{\alpha}}^k &H_{\mathbf{\rho},\mathbf{\rho}}^k
\end{array}\right] ; H_{i,j}^k = J^{k^T}_{i}J^{k}_{j}
\end{align}
Its important to recall that, due to the fact that each residual depends only on one pose and edgepoint, the Hessian has a sparse structure were $H_{\mathbf{\rho},\mathbf{\rho}}^k$, $H_{\mathbf{\rho},\mathbf{\alpha}}^k$ and $H_{\mathbf{\alpha},\mathbf{\alpha}}^k $ are diagonal, and $H_{\mathbf{\chi},\mathbf{\chi}}^k$ is block diagonal. 

In addition to this, variables $\mathbf{\alpha}^k$ and $\mathbf{\rho}^k$ only appear in sub-functional $E_k$. Thus, given an incremental update on the relative transformation $\mathbf{\delta \chi}^k$, is possible to solve for the optimal values of $\mathbf{\delta \alpha}^k$ and $\mathbf{\delta \rho}^k$ by taking the Schur Complement:
\begin{align}
\left[\begin{array}{c}   \mathbf{\delta \alpha}^{k} \\ \mathbf{\delta \rho}^{k} \end{array}\right]&=-
S_{\mathbf{\alpha}\mathbf{\alpha}\mathbf{\rho}\mathbf{\rho}} 
\left(
\left[\begin{array}{c} \mathbf{g}^{k}_\mathbf{\alpha}  \\ \mathbf{g}^{k}_\mathbf{\rho} \end{array}\right]
+ \left[\begin{array}{c}H_{\mathbf{\alpha},\mathbf{\chi}}^k \\H_{\mathbf{\rho},\mathbf{\chi}}^k \end{array}\right]  \mathbf{\delta \chi}^{k}\right)\label{ec:shur1}
\\
S_{\mathbf{\alpha}\mathbf{\alpha}\mathbf{\rho}\mathbf{\rho}} 
&=
\left[\begin{array}{cc} 
H_{\mathbf{\alpha},\mathbf{\alpha}}^k &H_{\mathbf{\alpha},\mathbf{\rho}}^k\\
H_{\mathbf{\rho},\mathbf{\alpha}}^k &H_{\mathbf{\rho},\mathbf{\rho}}^k
\end{array}\right]^{-1}
\end{align}
Which can be replaced into equation~\ref{ec:shur0} to obtain a form of the functional where edgepoint variables has been marginalized:
\begin{align}
E^l_k \left(\mathbf{\delta \chi}^k\right) &\simeq 
E^{l0}_k+ 2 y^{k^T}_{ \mathbf{\chi}\mathbf{\chi}} \mathbf{\delta \chi}^k +
\mathbf{\delta \chi}^{k^T} Y^k_{ \mathbf{\chi}\mathbf{\chi}}
\mathbf{\delta \chi}^k
\label{ec:shur3}
\end{align}
Where $ y^{k^T}_{ \mathbf{\chi}\mathbf{\chi}}$ and $Y^k_{ \mathbf{\chi}\mathbf{\chi}}$ stand for information vector and matrix respectively:
\begin{align}
y^k_{ \mathbf{\chi}}&=\mathbf{g}^{k}_\mathbf{\chi} -
\left[\begin{array}{cc} H_{\mathbf{\chi},\mathbf{\alpha}}^k&H_{\mathbf{\chi},\mathbf{\rho}}^k\end{array}\right]
S_{\mathbf{\alpha}\mathbf{\alpha}\mathbf{\rho}\mathbf{\rho}}
\left[\begin{array}{c} \mathbf{g}^{k}_\mathbf{\alpha}  \\ \mathbf{g}^{k}_\mathbf{\rho}  \end{array}\right]\\%
Y^k_{ \mathbf{\chi}\mathbf{\chi}}&= H_{\mathbf{\chi},\mathbf{\chi}}^k -  \left[\begin{array}{cc} H_{\mathbf{\chi},\mathbf{\alpha}}^k&H_{\mathbf{\chi},\mathbf{\rho}}^k\end{array}\right]
S_{\mathbf{\alpha}\mathbf{\alpha}\mathbf{\rho}\mathbf{\rho}}
\left[\begin{array}{c} H_{\mathbf{\chi},\mathbf{\alpha}}^{k^T}\\H_{\mathbf{\chi},\mathbf{\rho}}^{k^T}\end{array}\right]
\label{ec:YxxDef}
\end{align}
It these equations $\mathbf{\delta \chi}^k$ is a vector that stacks all transformation increments from PKF $k$ to all keyframes in its connectivity window $S^k$:
\begin{align*}
\mathbf{\delta \chi}^k &= \left[\begin{array}{ccc}
\mathbf{\chi}^{k-1}_k \dots \mathbf{\chi}^{k-S_k}_k \end{array}\right]
\end{align*}
\subsection{Joint Odometry Pose Graph Optimization}
\begin{figure*}[ht!]
	\centering
	\resizebox{0.9\textwidth}{!}{\includegraphics{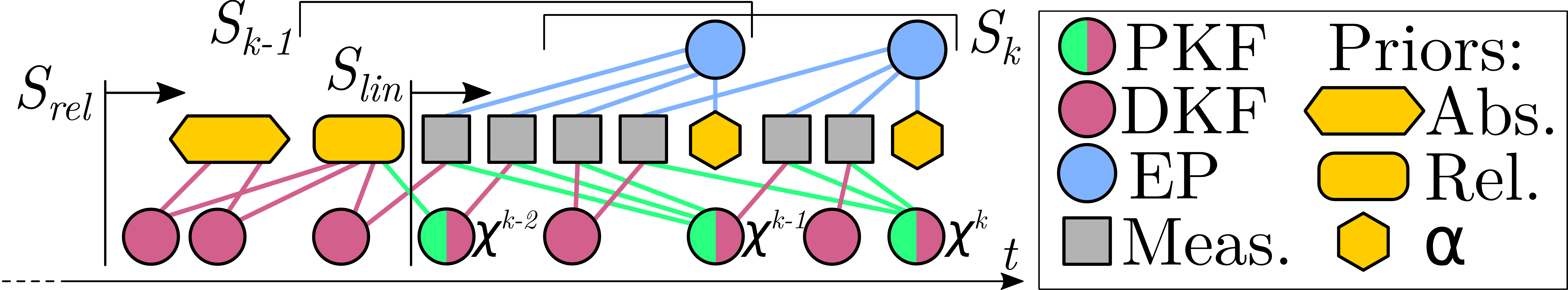}}
	\caption{Factor Graph of the optimization process showing variable interconnection and different optimization windows involved.}
	\label{fig:factor_grapth}
\end{figure*}
Given a set of sub-functionals $E_k$ approximated by equation \ref{ec:shur3},the overall optimization problem takes the form:
\begin{align}
E^{pg}(\mathbf{\chi}_{j,k})&=\sum_k 
2 y^{k^T}_{ \mathbf{\chi}\mathbf{\chi}}\mathbf{\delta \chi}^k + 
\mathbf{\delta \chi}^{k^T} Y^k_{ \mathbf{\chi}\mathbf{\chi}}
\mathbf{\delta \chi}^k 
\label{ec:pose_graph}
\\
\mathbf{\delta \chi}^k_j(\mathbf{\chi}_{j}, \mathbf{\chi}_{k})&= \log_{SE(3)} \left[\mathbf{\chi}_{j}^{-1} \mathbf{\chi}_{k} \mathbf{\bar{\chi}}^{k^{-1}}_{j}\right]\label{ec:d_pose}
\end{align}

Where constant terms has been dropped and $\mathbf{\delta \chi}^k_j$ is the tangent space distance from the current relative pose $\mathbf{\chi}_{j}^{-1} \mathbf{\chi}_{k}$ to the prior linearization point $\mathbf{\bar{\chi}}^{k}_{j}$. 

In order to solve this system using Gauss-Newton\cite{Absil2009pup}, a linear approximation of equation \ref{ec:d_pose} is used:
\begin{align}
\mathbf{\delta \chi}^k_j(\mathbf{\chi}_{j}, \mathbf{\chi}_{k})&\simeq \log_{SE(3)} \left[\mathbf{\bar{\chi}}_{j}^{-1} \mathbf{\bar{\chi}}_{k} \mathbf{\bar{\chi}}^{k^{-1}}_{j}\right] + J^\delta_k \mathbf{\delta\chi}_{k} + J^\delta_j \mathbf{\delta\chi}_{j}\label{ec:delta_rel}
\end{align}

It should be noted that, while absolute poses $\mathbf{\chi}_{j,k}$ are iteratively re-linearized in every optimization step, sub-functional energy priors $E_k$ are selectively re-linearized only when needed \cite{Kaess2008TOR}, \cite{Kaess2012ijrr} (see sec. \ref{sec:ep_update}). Therefore it may happen that $\mathbf{\bar{\chi}}_{j}^{-1} \mathbf{\bar{\chi}}_{k} \ne \mathbf{\bar{\chi}}^{k}_{j}$.

The Jacobians are defined with respect to an infinitesimal increment on SE3 :

\begin{align}
 J^\delta_k &= \left. \frac{\partial \log_{SE(3)}\left[
 	\mathbf{\bar{\chi}}_{j}^{-1}\exp_{SE(3)}[\mathbf{\delta\chi}_{k}] \mathbf{\bar{\chi}}_{k} \mathbf{\bar{\chi}}^{k^{-1}}_{j} \right]}
 {\partial \mathbf{\delta\chi}_{k}}\right|_{\mathbf{\delta\chi}_{k}=0} \label{eq:jac_j1} \\
 J^\delta_j &= \left. \frac{\partial \log_{SE(3)}\left[
 	\mathbf{\bar{\chi}}_{j}^{-1}\exp_{SE(3)}[\mathbf{\delta\chi}_{j}]^{-1} \mathbf{\bar{\chi}}_{k} \mathbf{\bar{\chi}}^{k^{-1}}_{j} \right]}
 {\partial \mathbf{\delta\chi}_{j}}\right|_{\mathbf{\delta\chi}_{j}=0} \label{eq:jac_j}
\end{align}

Both jacobians are similar, in fact noting that $\exp_{SE(3)}[\mathbf{\delta\chi}_{j}]^{-1} = \exp_{SE(3)}[-\mathbf{\delta\chi}_{j}]$ it can be shown that $J^\delta_j = -J^\delta_k$.

Using the SE3 adjoint operator, its possible to ``shift'' $\exp_{SE(3)}[-\mathbf{\delta\chi}_{k}]$ to the left of $\mathbf{\delta\chi}_{k}$:

\begin{align}
J^\delta_k &= \left. \frac{\partial \log_{SE(3)}\left[
	\exp_{SE(3)}[\mathbf{\delta\chi}] \mathbf{\bar{\chi}_e} \right]} 
{\partial \mathbf{\delta\chi}}\right|_{\mathbf{\delta\chi}=0} \mathcal{ADJ}_{SE3} [\mathbf{\bar{\chi}}_{j}^{-1}]\label{eq:adjoint}
\end{align}

In this equation $\mathbf{\bar{\chi}_e} = \mathbf{\bar{\chi}}_{j}^{-1} \mathbf{\bar{\chi}}_{k} \mathbf{\bar{\chi}}^{k^{-1}}_{j}$ is the error transform. Its twist vector is defined in terms of a translational ($\mathbf{\bar{v}_e}$) and rotational component ($\mathbf{\bar{\omega}_e}$):

\begin{align}
 \mathbf{\bar{\chi}_e}
&= \exp_{SE(3)}\left[\begin{array}{c}
\mathbf{\bar{v}_e} \\
\mathbf{\bar{\omega}_e}
\end{array}\right]
\end{align} 

Analytical expressions for the derivative in equation~\ref{eq:adjoint} are complex, but they can be found in \cite{EadeLieDeriv}, \cite{Barfoot2017CUP} and in the source code of this work.

It is interesting to note that, in the case were sub-functional $E_k$ has been marginalized in the current optimization step, $\mathbf{\bar{\chi}_e}$ equals identity and the jacobian becomes the adjoint:
\begin{align}
\left. J^\delta_k \right|_{ \mathbf{\bar{\chi}}^{k}_{j} = \mathbf{\bar{\chi}}_{j}^{-1} \mathbf{\bar{\chi}}_{k}} &=\mathcal{ADJ}_{SE3} [\mathbf{\bar{\chi}}_{j}^{-1}]\\
&= \left[\begin{array}{cc}
R_{j}^{T}&[-R_{j}^{T}\mathbf{t}_{j}]_\times R_{j}^{T}\\
0&R_{j}^{T}
\end{array}\right]
\end{align} 
Equation \ref{ec:delta_rel} can be plugged into \ref{ec:pose_graph} to obtain a linear system of equations on the increments in absolute poses. This is a sparce and mainly block diagonal system (except for the connections generated by the loop closure), with non zero entries only in poses that share a mutually observed landmarks. 

Solutions are obtained using Sparce Cholesky Decomposition and, once increments are obtained, absolute pose linearization points are updated:
\begin{align}
\mathbf{\chi}_{j,k} = \mathbf{\delta\chi}_{j,k} \boxplus 
\mathbf{\bar{\chi}}_{j,k} && \forall k,j 
\end{align}
\subsection{Edgepoints Update}
\label{sec:ep_update}
As already mentioned, not all PKFs functionals are re-linearized in every optimization step, provided that computing residuals, Jacobians and Schur complements in equations \ref{eq:rep_error_rel_2} to \ref{ec:shur3} involves calculations with every edgepoint. Given that in average a PKF defines 5000-12000 edgepoints, these operations comprise most of the optimization algorithm running time. 

On top of this, once a PKF has been re-linearized and optimized many times and falls outside the co-visibility window, effective update in variables become negligible and equation \ref{ec:shur3} starts being a good approximation to the full functional~\cite{Kaess2008TOR}. This way, sub-functionals are re-linearized only when they are within a certain temporal window and a significant pose change occurred during the last optimization step. This is measured by comparing the distance from the current relative transformations involving the sub-functional to its last linearization point, relative to the later:
\begin{align}
\max_j\left(\frac{|\mathbf{\chi}_{j}^{-1} \mathbf{\chi}_{k} \mathbf{\bar{\chi}}^{k^{-1}}_{j}|}{|\mathbf{\bar{\chi}}^k_{j}|}\right)<\epsilon
\label{eq:relin}
\end{align} 
For the PKFs that has been re-linearized, edgepoint's variables $\rho$ and $\alpha$ are updated using equations \ref{ec:shur1}. Relative increments are first obtained with equation~\ref{ec:delta_rel} and Jacobians \ref{eq:jac_j} and \ref{eq:jac_j1}, which are cached from the PGO.

Because only edgepoints belonging to re-linearized PKFs are updated, the constant term in expression~\ref{ec:delta_rel} vanishes as $\mathbf{\bar{\chi}}_{j}^{-1} \mathbf{\bar{\chi}}_{k} = \mathbf{\bar{\chi}}^{k}_{j}$. Also the derivative in equation \ref{eq:adjoint} falls back to identity, leaving only the adjoint.

\subsection{Relative vs. Absolute coordinates in local BA}
\label{sec:rel_coords}
In a full BA scheme every PKF sub-functional would be re-linearized in each step. In this case relative formulation in $E_k$ serves only as an intermediate step, yielding an equivalent solution to formulating equation \ref{eq:rep_error_rel} in absolute coordinates (due to Jacobians chain rule). This is to be expected as linearization should be independent of the intermediate variables used to obtain it.

The case is substantially different when re-linarization is avoided and \ref{ec:shur3} is used as an approximation to the true functional, in contrast to a one constructed using absolute coordinates.
 
Formulating relative coordinates priors has interesting advantages \cite{Sibley2009tr}. To begin with, BA  classical structure where each factor involves a landmark and only one pose is maintained; thus simplifying Schur Complement and Hessian computation. But more importantly, the relative formulation is a better approximation of the real functional. 

A relative parametrization naturally expresses the invariance to a change in absolute pose, and marginalized priors have a scale invariance form due to a rank deficiency (gauge freedom). 
In fact, the functional will not penalize increments in the direction of the translation components of the linearization point:
\begin{align}
\mathbf{\delta \chi}^{k0} &= \left[\begin{array}{ccc}
\left[\begin{array}{cc}\mathbf{\bar{t}}^{k-1}_k \mathbf{0}\end{array}\right]
 & \dots &
\left[\begin{array}{cc}\mathbf{\bar{t}}^{k-S_k}_k \mathbf{0} \end{array}\right] 
\end{array}\right]
\\
E^l_k \left(\lambda \mathbf{\delta \chi}^{k0}\right) &=
E^l_k \left(\mathbf{0}\right) = E^{l0}_k \quad  \forall \lambda \label{eq:marg_inv}
\end{align}
This implies that the functional will not penalize translation expansions or shrinks as long as they belong to the direction pointed by the linearization point, which after a few iterations should converge to the optimal. This interesting property allows modeling scale drift without the use of SIM3 \cite{Strasdat2010RSS} optimization. A working example of this idea can be seen in Figure\ref{fig:scale_drift}.

A proof of equation \ref{eq:marg_inv} can be sketched by observing that the linearized residual \ref{eq:rep_error_rel_2}:
\begin{align*}
r^{k,j}_i&=\bar{r}^{k,j}_i + 
J^{k,j}_{i,\alpha}  \delta \alpha_i +
J^{k,j}_{i,\rho}  \delta \rho_i +
J^{k,j}_{i,\mathbf{v}_k^j} \mathbf{\delta v}^j_k +
J^{k,j}_{i,\mathbf{\omega}_k^j} \mathbf{\delta \omega}^j_k 
\end{align*}

is invariant to an increment $\lambda \mathbf{\delta \chi}^{k0}$ if $\delta \rho_i =-\lambda \bar{\rho}_i$ and $\delta \alpha_i =0$. Because marginalized functional is the result of a minimization over $\delta \rho_i$ and $\delta \alpha_i$ the following holds:

\begin{align}
E^l_k \left(\lambda \mathbf{\delta \chi}^{k0}\right)
&\leq E^{0}_k \quad  \forall \lambda
\\
E^{l0}_k+ 2 y^{k^T}_{ \mathbf{\chi}\mathbf{\chi}} \mathbf{\delta \chi}^{k0} \lambda+
\lambda\mathbf{\delta \chi}^{{k0}^T} Y^k_{ \mathbf{\chi}\mathbf{\chi}}
\mathbf{\delta \chi}^{k0}\lambda
&\leq E^{0}_k \quad  \forall \lambda
\\
E^{l0}_k+ \beta_1 \lambda + \beta_2 \lambda^2
&\leq E^{0}_k \quad  \forall \lambda \label{eq:beta}
\end{align}

In order for \ref{eq:beta} to be true for every $\lambda$ it must be $\beta_1=\beta_2=0$, implying invariance to a change $\lambda \mathbf{\delta \chi}^{k0}$.
\subsection{Pose Marginalization}
\label{sec:pose_marg}
Problem~\ref{ec:pose_graph} involves poses only, therefore is much faster to solve than full BA. However, it grows linearly with time, becoming computationally expensive. To keep time constrained, marginalization is applied to old enough poses, thus not changing any more \cite{Leutenegger2015ijjr}. A wider marginalization window $S_{rel}$ is defined, such that relative PKF priors falling outside this window are fused into a single absolute prior.

This procedure is simple, the system creates an absolute position prior to fix both position and scale to its initial condition. Absolute priors involve a vector of absolute correlated poses $\mathbf{\chi}^{p}$ and take the following generic form:
\begin{align}
E_{p} = \left(\mathbf{\chi}^{p} \boxminus \mathbf{\bar{\chi}}^{p} \right)^T Y^p_{\mathbf{\chi}\mathbf{\chi}} \left( \mathbf{\chi}^{p} \boxminus \mathbf{\bar{\chi}}^{p} \right)
\end{align}
The absolute prior is jointly optimized with the rest of problem \ref{ec:pose_graph} until a PKF gets a link to a keyframe outside the marginalization window. In this case, its corresponding sub-functional $E_k$ is jointly optimized with the absolute position prior, updating the later. Finally, the absolute prior is constrained using Schur marginalization to have all its measurements inside the marginalization window $S_{rel}$. Needless to say, this is a sliding window, therefore old PKFs are incrementally merged in the absolute prior.

A factor graph depicting the overall structure of the problem and different windows involved is shown in Figure~\ref{fig:factor_grapth}. Marginalized relative priors are mantained in memory in case they need to be recovered in the future for full PGO (refer to section \ref{sec:loop_optim}).

\subsection{Overall Algorithm}

The optimization algorithm can be summarized in the following steps (performed for each iteration): 
\begin{enumerate}
	\item Check which PKFs should be re-linearized (\ref{sec:ep_update}, equation~\ref{eq:relin}). If none, go to step 3.
	\item Re-linearize needed PKFs using equations \ref{eq:rep_error_rel_2} to \ref{ec:YxxDef}.
	\item Marginalize old poses and re-build absolute prior according to section \ref{sec:pose_marg}.
	\item Solve a PGO problem on absolute pose variables and update them.
	\item Update landmarks using equations \ref{ec:shur1}, only for PKFs re-linearized in this iteration .
\end{enumerate}
Before running the optimization, loop closure check is performed if the current frame is a PKF. If a match is found, a slightly different optimization algorithm is run (see section~\ref{sec:loop_optim}).

\section{Loop Closures}
\label{sec:loop_closure}

Loop closure is a key aspect of SLAM systems, especially monocular ones, that greatly helps to keep the overall error constrained. However, this is a challenging problem on its own and is out of the scope this paper which focuses on edgemaps association and optimization. Therefore, inspired in similar works \cite{Mur2017tor,Gao2018iros}, well established feature based methods for loop candidate search and pre-alignemnt have been used, in order to show SE-SLAM capabilities apart from the loop closure problem itself. Leaving a full edge-based loop closing mechanism proposed for future work.

\subsection{Loop Candidates Search}
In order to find possible loop candidates, only for PKFs ORB features are extracted \cite{Rublee2011iccv} and uses to compute a binary representation using a Bag of Words (BoW) dictionary implementation \cite{Galvez2012tog}, thus creating a BoW database. Every time a new PKF is created, it is compared with all previous ones outside its covisibility window, obtaining a similarity score. The 2 PKF with higher score above the overall average are marked as potential loop candidates.

Candidate's keypoints are matched with their descriptors and a transformation is computed by solving the PnP problem with RANSAC using OpenCV. To that end, candidate's keypoints depth is estimated by averaging neighbouring edgepoints inverse depth on a window around each point. This transformation is further defined using the edge tracking method described in section~\ref{sec:tracking}.

To check a candidate's validity, its edgepoints are reprojected to the current frame. If sufficient edgepoints present low reprojection error and are inside the current frame's FOV, the candidate is accepted. 

\begin{figure}[ht]
	\centering
	\resizebox{0.9\columnwidth}{!}{\includegraphics{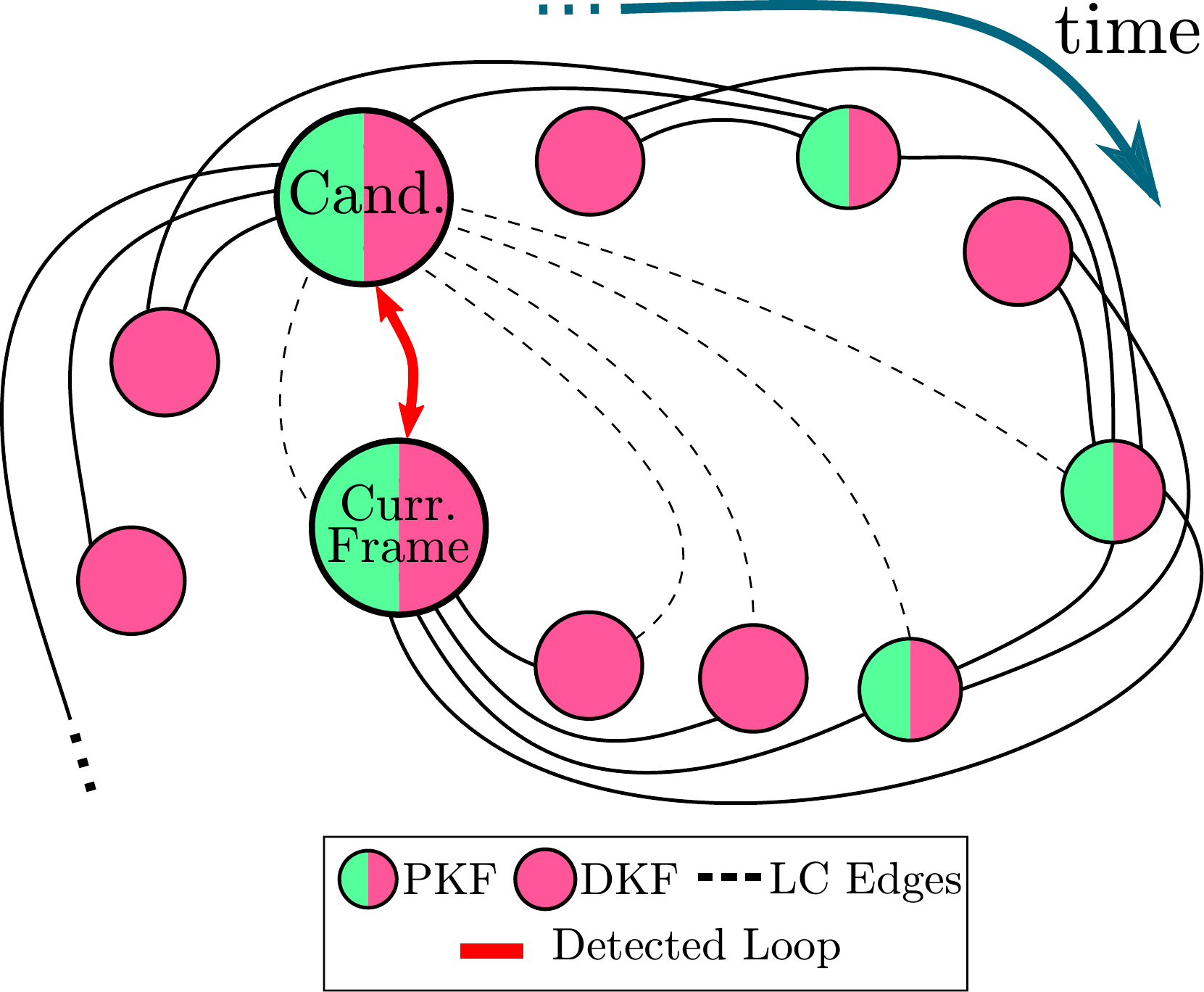}}
	\caption{Loop closure scheme showing the edges added to the pose graph (dashed) from the accepted loop candidate (solid red) to current PKF's co-visibility window. These mutual connections anchor depth and scale.}
	\label{fig:loop_closure}
\end{figure}

\subsection{Optimization}
\label{sec:loop_optim}
Once a candidate has been aligned and associated with the current PKF, its data association is used to propagate matches from the candidate PKF to every keyframe in the current keyframe's covisibility window. This is done simply by connecting successive matches.

Once this new connections are established, optimal transformation from candidate PKF to each of these keyframes is obtained by minimizing the reproyection error~\ref{eq:rep_error_rel} with respect to the relative pose only.

Next, a joint relative odometry prior $E_{cand}$ from the candidate to keyframes in both candidate and current covisivility windows, is built using the procedure described in section~\ref{sec:schur}. This prior is constructed using the optimized poses obtained in the previous step as linearization point, which gives a good hint for the true optimal point.


Once this prior is computed, all relative priors are re-activated and the pose graph is optimized completely (marginal absolute prior is discarded). Active covisibility window edgepoints are also included in the optimization. The pose graph structure in this case is illustrated in Figure~\ref{fig:loop_closure}.

Usually a few iterations ($\sim 3$) are required for convergence. Upon this, all relative priors falling outside the relative marginalization window $S_{rel}$ are fused again into a single absolute marginal prior. The system then resumes the process described in previous sections.

It is paramount to interpret the prior $E_{cand}$ as a compressed form of the reprojection error functions from the candidate PKF's edgepoints to every keyframe it is connected to, both in the current and its own covisibility windows. As shown in Figure~\ref{fig:loop_closure}.

Given that such connections involve more that one mutual pose, not only position but also scale is connected. When PGO is done with all relative priors, together with the active edgepoints, scale adapts to the one given by the candidate PKF, that is connected through is covisibility window to older measurements, as is shown experimentally in Figure~\ref{fig:scale_drift}. In this setup, scale freedom is not given by SIM3 optimization \cite{Strasdat2010RSS} but by the rank deficiency of the joint odometry priors (section \ref{sec:rel_coords}). 

\section{Implementation Details}
\label{sec:imp_details}
\subsection{Edge Detection}

Up to this point, little has been said about the edge extraction procedure. This account for the fact that presented algorithm is intended to be detection method agnostic. However, as edges are the presented algorithm's input, it is expected for different methods to have a significant impact in its performance.

The edge detection algorithm is based on maximal gradient on a smoothed image \cite{Canny1987rcv}. Despite being simple, it yields well localized edges, which is the most desirable feature for a SLAM algorithm, together with repeatability. 

Non Maximal suppression and sub-pixel position is estimated for each edgepoint and third derivative threshold is also applied \cite{Lindeberg1999ap}, as this helps discriminate continuous gradients.

Finally, the edgemap is optionally decimated by half to reduce the overall number of variables. This helps to increase performance in terms of running speed, by eliminating redundant variables, while keeping full resolution position and connectivity.

\subsection{Optimization Parallelization}

Most computation time of the algorithm is spent in the edgepoints marginalization step (section~\ref{sec:schur}). Marginalizations of each PKF prior are independent of each other, as their information is connected in the PGO. Therefore, these calculations are parallelized using a thread pool of 8 threads. 

Finally, secondary sparseness is also exploted inside each prior calculation by separating edgepoints in groups according to how far they expand in the covisibility window. Schur complement can be applied separately in each group, adding the results in a later stage. These groups are also parallelized. 

\section{Evaluation}
\label{sec:results}
\begin{figure*}[ht!]
	\centering
	\resizebox{1\textwidth}{!}{\includegraphics{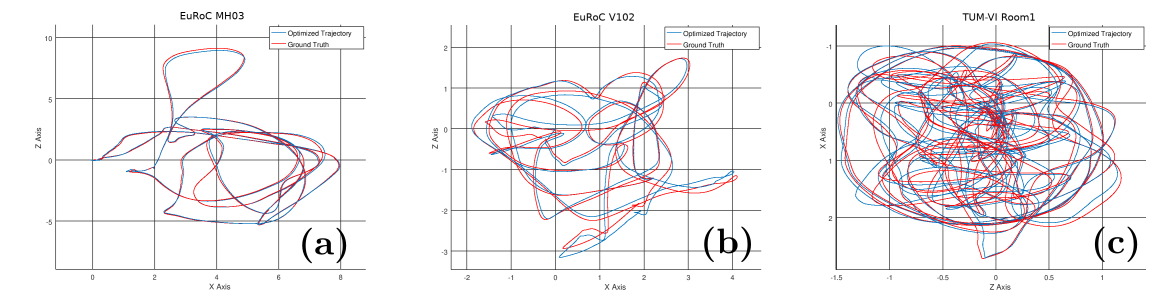}}
	\caption{Top down views of fully optimized trajectories compared to provided groun-truth. This show SE-SLAM capabilities to estimate trajectories in both large (a) and small (b,c) environments and keep consistency after several loops (c)}
	\label{fig:trays}
\end{figure*}
The presented system was benchmarked with the EuRoC dataset\cite{Burri2016ijjr}, which is widely used for evaluation of SLAM methods, indoor sequences from the TUMVI dataset \cite{Schubert2018IROS} and the ICL-NUIM dataset \cite{Handa2014ICRA}. Round Mean Squared Absolute Trajectory Error (ATE) is the chosen metric for evaluation, as it provides an overall benchmark of the position accuracy.
Due to the monocular nature of the algorithm, SIM3 alignment of the trajectory with respect to the provided ground-truth and ATE error calculation were performed using a public algorithm \cite{MurGit}, based on the method proposed in \cite{Sturm2011rss} and \cite{Umeyama1991PAMI}.

The algorithm was run on a desktop computer featuring an Intel(R) Core(TM) i9-9900K CPU @ 3.60GHz with 64GB of RAM, however the system's memory footprint does not exceed 2GB. The system is initialized using \cite{Jose2015iccv}, engaging full optimization once a few frames have been accumulated. Similar to other methods \cite{Engel2014eccv}, the first and last parts of the datasets, were the camera is still or shook for IMU initialization, are stripped off as it compromises monocular initialization. 

Two versions of the method are tested, namely SESLAM-1 and SESLAM-3 which perform one and three optimization iterations per frame respectively.

\subsection{Results on EuRoC Dataset}

EuRoC-MAV is a well established state of the art benchmark for SLAM systems and is the main dataset used for evaluating this algorithm. It provides two sets of realistic indoors sequences taken from a quadrotor MAV, rated by its difficulty. The first set features five sequences of a large machine hall with different degrees of dynamic motions and illumination changes. The second set was taken inside two rooms equipped with a VICON tracking system and while less space is covered, they are more challenging in terms high dynamic motions. This whole dataset is targeted for stereo visual inertial algorithms, hence running pure monocular systems introduces additional difficulties, specially in pure camera rotation cases, where the whole edgemap is renewed without motion.

Table~\ref{tbl:errors_vicon} show ATE for the Machine Hall and Vicon Room sequences respectively, of SE-SLAM compared against two widely known algorithms: OBSLAM2 \cite{Mur2017tor} representing the feature based SLAM systems and L-DSO \cite{Gao2018iros} for direct ones. Both methods are state of the art in terms of Monocular SLAM and were run using configurations provided by the authors.
\begin{table}[ht!]
	\centering
	\begin{tabular}{@{}lcccc@{}}
		\hline\noalign{\smallskip}
		& SESLAM& ORBSLAM2 & LDSO\\
		\noalign{\smallskip}\hline\noalign{\smallskip}
		MH01&$6.8\pm0.2$ ($80\%$)    &$4.6\pm0.2$  ($100\%$) &$5.0\pm0.5$ ($100\%$)\\
		MH02&$8.7\pm3.7$ ($100\%$)   &$3.6\pm0.1$  ($100\%$) &$5.3\pm0.5$ ($100\%$)\\
		MH03&$8.2\pm1.2$ ($40\%$)    &$3.9\pm0.1$  ($100\%$) &$8.6\pm0.1$ ($80\%$)\\
		MH04&- ($0\%$)               &$9.4\pm3.5$  ($100\%$) &$9.5\pm1.6$ ($100\%$)\\
		MH05&- ($0\%$)               &$5.2\pm0.3$  ($100\%$) &$7.6\pm0.6$ ($100\%$)\\
		\noalign{\smallskip}\hline\noalign{\smallskip}
		VR1\_1&$8.6\pm0.1$ ($60\%$)  &$9.6\pm0.1$  ($100\%$) &$10.0\pm0.6$ ($100\%$)\\
		VR2\_1&$9.6\pm0.2$ ($100\%$) &$8.0\pm0.1$  ($100\%$) &$9.0\pm0.2$ ($100\%$)\\
		VR1\_2&$14.7\pm4.2$ ($80\%$) &$10.3\pm0.1$  ($100\%$) &$20.5\pm13.8$ ($60\%$)\\
		VR2\_2&- ($0\%$)             &$17.2\pm0.3$  ($100\%$) &$20.0\pm0.5$  ($100\%$)\\
		\noalign{\smallskip}
	\end{tabular}
	\caption{Global ATE [m] comparison against monocular SLAM systems on EuRoC Machine Hall (MH) and Vicon Room (VR) sequences. F = failed ($ATE > 0.5m$).} 
	\label{tbl:errors_vicon}
\end{table}
It can be seen that SE-SLAM presents similar accuracy to that of both methods and in particular SESLAM-3 only introduces a significant improvement in difficult sequences, where high dynamic movements introduce large amounts of new data per frame. The cause of failure in the most challenging datasets is loss of tracking due to strong intensity changes and large motions with image blur. The system does not recover from this cases as it lacks a re-localization mechanism, which is proposed as future work


It is important to highlight that both ORB-SLAM2 and LDSO are the result of long incremental work in their respective fields, in contrast to edge-based SLAM. In light of SE-SLAM competitive results in these sequences, it is clear that this approach is very promising and has room for improvements that could be carried out in the future.

Top down views of resulting trajectories for sequences MH03 and V102 are shown in fig. \ref{fig:trays}(a,b).

\subsection{Results on TUMVI Dataset}

TUMVI Dataset is a newer than the former, comprising many sets of indoor and outdoor sequences. It is a challenging dataset aimed to evaluate stereo visual inertial algorithms over long sequences with dynamic motions and illumination changes. The Room sequences were chosen for evaluation, as they are the only ones with ground truth data for whole sequences.
\begin{table}[ht!]
	\centering
	\renewcommand{\tabcolsep}{5pt}
	\begin{tabular}{@{}lccccc@{}}
		\hline\noalign{\smallskip}
		& SESLAM-1& OKVIS \cite{Leutenegger2015ijjr} & ROVIO\cite{Bloesch2017ijrr} & VINS \cite{Qin2018TOR}\\
		\noalign{\smallskip}\hline\noalign{\smallskip}
		Room 1&$0.072\pm0.003$ ($80\%$) &0.06&0.16&0.07\\
		Room 2&$0.096\pm0.011$ ($100\%$) &0.11&0.33&0.07\\
		Room 3&- ($0\%$) &0.07&0.15&0.11\\
		Room 4&- ($0\%$) &0.03&0.09&0.04\\
		Room 5&$0.085\pm0.015$ ($100\%$) &0.07&0.12&0.20\\
		Room 6&$0.070\pm0.002$ ($100\%$) &0.04&0.05&0.08\\
		\noalign{\smallskip}
	\end{tabular}
	\caption{Global ATE [m] comparison against Visual Inertial SLAM systems on TUMVI Room sequences. F = failed ($ATE > 0.5m$).} 
	\label{tbl:errors_tum}
\end{table}
Test results are shown in Table~\ref{tbl:errors_tum}, where SE-SLAM was compared with the results published together with the dataset \cite{Schubert2018IROS}, to ensure the other systems have configuration intended by the authors. It can be seen again that even though SE-SLAM is a monocular system, it produces competitive results even against the state of the art visual-inertial SLAM systems.

The fully optimized trajectory for sequences Room1 is depicted in fig. \ref{fig:trays}(c).

\subsection{Results on ICL-NUIM Dataset}
Finally, SE-SLAM was also run in the Office4 sequences of ICL-NUIM synthetic dataset, in order to compare it with the edge-based algorithm presented in \cite{Maity2017iccv}. Results are shown in Table~\ref{tbl:errors_icl}, where ATE was taken from \cite{Maity2017iccv}, as they haven't released their code.

It should be noted that the ICL is a dataset targeted for RGBD SLAM and reconstruction systems. The office trajectories employed by \cite{Maity2017iccv} are short sequences with slow motions. Given that roughly 20\% of the sequence is spent on initialization, the authors consider these to be a sub-optimal way to assess this type of SLAM systems. Nevertheless, our system outperforms \cite{Maity2017iccv}, except for a case were a tracking failure occurs.
\begin{table}[ht!]
	\centering
	\begin{tabular}{@{}lcc|lcc@{}}
		\hline\noalign{\smallskip}
		Seq & SESLAM-1& \cite{Maity2017iccv} & Seq & SESLAM-1& \cite{Maity2017iccv}  \\
		\noalign{\smallskip}\hline\noalign{\smallskip}
		ICL/Office 0&0.018&0.032 & ICL/Office 2&0.458&0.030\\
		ICL/Office 1&0.149&0.195 & ICL/Office 3&0.028&0.049\\
		\noalign{\smallskip}
	\end{tabular}
	\caption{Global ATE [m] comparison against Edge SLAM \cite{Maity2017iccv} on ICL-NUIM Office sequences.} 
	\label{tbl:errors_icl}
\end{table}
\subsection{Reconstruction Results}

\begin{figure}[t!]
	\centering
	\resizebox{0.9\columnwidth}{!}{\includegraphics{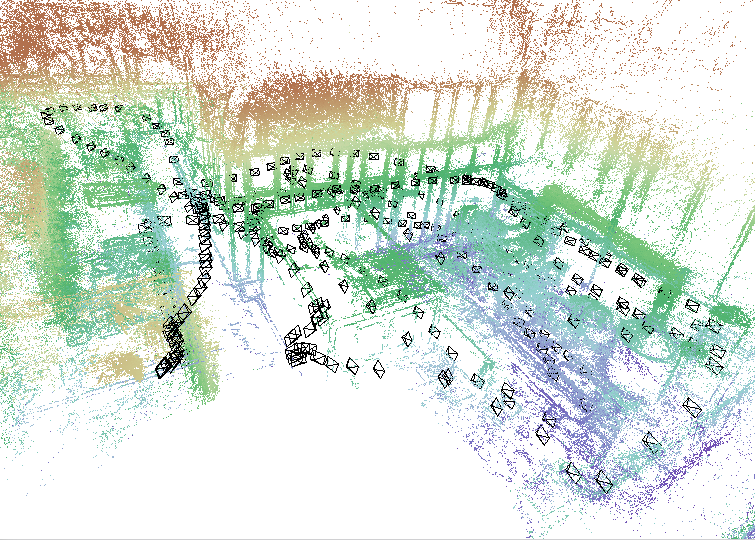}}
	\caption{Final edgemap and trajectory obtained in the Machine Hall MH03 sequence.}
	\label{fig:reconstruction2}
\end{figure}

\begin{figure}[t!]
	\centering
	\resizebox{0.9\columnwidth}{!}{\includegraphics{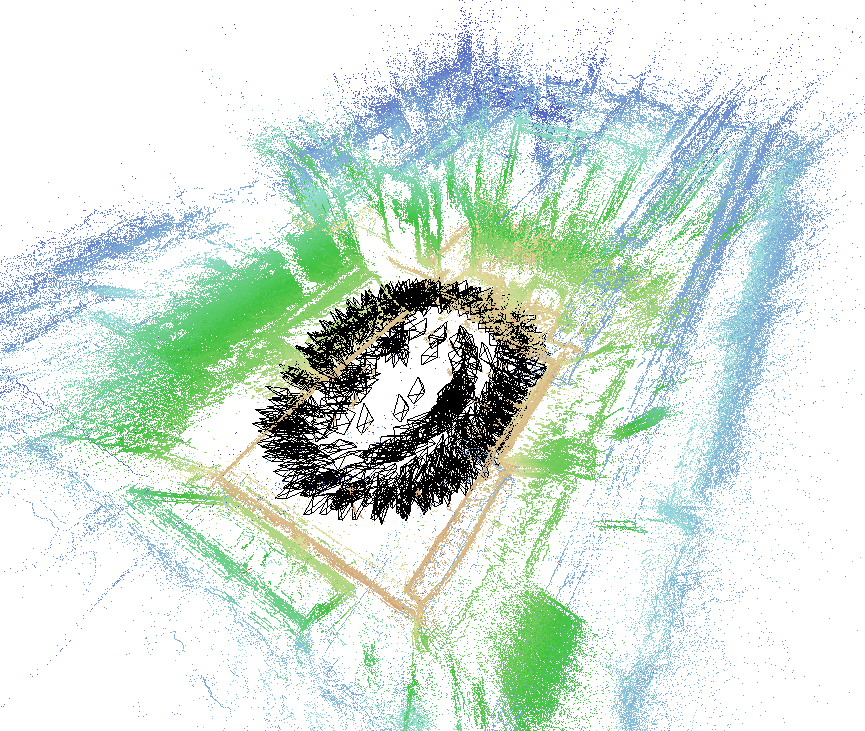}}
	\caption{Final edgemap and trajectory obtained in Room 2 sequence.}
	\label{fig:reconstruction3}
\end{figure}

Reconstruction output is a key aspect of this method. Examples are shown in Figures~\ref{fig:reconstruction},~\ref{fig:reconstruction2} and \ref{fig:reconstruction3}. The advantage in terms of structural information provided, with respect to sparse systems, is easy to see. Noteworthy is how the system keeps consistency in Room2 sequence over a 2.24 min span with several loops happening.

\subsection{Running times}
Given that SE-SLAM is a semi-dense method optimizing every significant edgepoint in images ($\sim$ 5K to 12K extracted per frame), the computational burden is higher compared to sparse methods. Also, running time heavily depends on each particular scene structure. Despite this, the system presents good performance in most datasets. This is due to its PKF selection policy, hierarchical optimization exploiting primary and secondary spareness and parallelization. Measured running times are reported in Table~\ref{tbl:rt}. 

\begin{table}[ht!]
	\centering
	\begin{tabular}{@{}lcc|lcc@{}}
		\hline\noalign{\smallskip}
		Seq& FPS& EP/F & Seq & FPS& EP/F \\
		\noalign{\smallskip}\hline\noalign{\smallskip}
		MH01&26&11102 & Room 1&21&9085\\
		MH02&28&10808 & Room 2&22&8806\\
		MH03&26&9422 & Room 3&14&947\\
		MH04&17*&7654 & Room 4&-&-\\
		MH05&16*&7751 & Room 5&19&8403\\
		V101&36&7628 & Room 6&29&8322\\
		V201&41&6948 & ICL-0 &26&11597\\
		V102&32&5651 & ICL-1&34&8832\\
		V202&17*&5974 & ICL-2 &27&11841   \\ 
		V103&-&- &   ICL-3& 28 &12186  \\ 
		V203&-&- & & &  \\ 
		\noalign{\smallskip}
	\end{tabular}
	\caption{Running times and Averaged extracted edgepoints, EP/F = Edgepoints per frame, FPS = mean frames per second. *SESLAM-3 results when SESLAM-1 fails ($ATE>0.5m$)} 
	\label{tbl:rt}
\end{table}
\subsection{Scale Drift and Loop Closure}

\begin{figure}[ht!]
	\centering
	\resizebox{0.9\columnwidth}{!}{\includegraphics{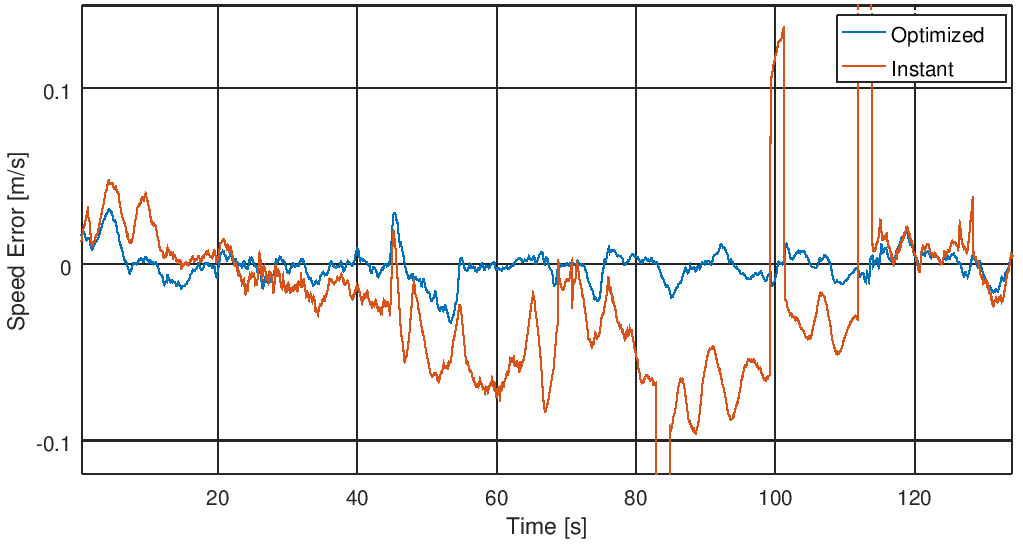}}
	\caption{Speed error with respect to ground-truth on MH01 sequence of incremental (instant) and fully optimized trajectories. Negative values imply underestimated scale}
	\label{fig:scale_drift}
\end{figure}

Finally, Figure~\ref{fig:scale_drift} shows the estimated speed error for both incremental and fully optimized trajectories on EuRoC MH01 sequence. In the incremental case the system underestimates speed as a consequence of increasing scale drift. When a loop closure occurs ($t\approxeq115s$), PGO performs a full trajectory optimization, converging to a consistent scale.

\section{Conclusion}

In this paper SE-SLAM, a novel semi-dense structured edge-based monocular SLAM system, that achieves competitive results in challenging datasets, compared to other state of the art systems was presented. 
The edge-based nature of the system inherently builds a semi-dense reconstruction of the environment, providing relevant structure data for further navigation algorithms, leveraging from most relevant image information even in textureless environments. 
An edge-based system is not trivial to build, mainly because edges are not easy to associate, track and optimize over time, as they lack descriptors and do not present biunivocal correspondence, unlike point features. These issues were tackled by developing methods to adapt SLAM theory to the edges nature, including:
an approach to match edges between frames and between relatively far keyframes in a consistent manner; 
a parametrization, residual and variable selection strategy that exploits edges' nature to make the optimization problem treatable, avoiding its rapid size increase; and a non-linear optimization that better adapts to the edge optimization problem. 

Finally, SE-SLAM was benchmarked in the widely used EuroCMAV, the newly TUMVI and the ICL-NUIM datasets. As can be seen in the results section, the system shows comparable precision to state of the art feature-based and dense/semi-dense systems, while achieving real-time operation in several cases, using edges as sole input in all the stages but loop closure.

Proposed future work includes extending the use of edges to the loop closure stage to make the system completely edge-based, trying edge sparsification and interpolation to reduce processing time and exploring different detectors and association strategies. Developing a re-localization stage is also proposed, so as to recover from tracking failures.

In light of SE-SLAM competitive results, it is clear that the edge-based approach to SLAM is very promising and has room for improvements to reach its full maturity. To encourage such developments, SE-SLAM source code will be soon released as open source.

\bibliographystyle{ieeetr}
\small
\bibliography{references}
\end{document}